\documentclass[12pt,letterpaper]{article}
\usepackage{authblk} 
\usepackage[comma,authoryear]{natbib} 
\usepackage{amssymb}
\usepackage{amsmath}
\usepackage{graphicx}
\usepackage{url}
\usepackage{array}
\usepackage{color,soul} 
\usepackage{float} 
\usepackage{hyperref}
\usepackage[colorinlistoftodos]{todonotes}
\usepackage{fancyhdr} 
\usepackage[colorinlistoftodos]{todonotes}
\usepackage{booktabs} 
\usepackage{rotating}
\usepackage{multirow}
\usepackage[multiple]{footmisc} 
\usepackage{geometry}
\usepackage{layout}

\usepackage{tabu}
\usepackage{booktabs}
\def \Npatients {82}
\def \pconvnet {$P{-}\mathrm{ConvNet}$}
\def \rconvnet {$R{-}\mathrm{ConvNet}$}
\def \roneconvnet {$R_1{-}\mathrm{ConvNet}$}
\def \rtwoconvnet {$R_2{-}\mathrm{ConvNet}$}

\def \testmean {71.8}
\def \teststd {10.7}
\def \testmin {25.0}
\def \testmax {86.9}
\def \testperf {\testmean{}$\pm$\teststd{}\%} 
\def \trainmean {83.6}
\def \trainstd {6.3}
\def \trainperf {\trainmean{}$\pm$\trainstd{}\%}

\newcommand\figscale{0.99} 
\begin{document}
\clearpage
\newpage 
\title{DeepOrgan: Multi-level Deep Convolutional Networks for Automated Pancreas Segmentation}
\author[1]{\small Holger R. Roth\thanks{holger.roth@nih.gov, h.roth@ucl.ac.uk}}
\author[1]{\small Le Lu}
\author[1]{\small Amal Farag}
\author[1]{\small Hoo-Chang Shin}
\author[1]{\small Jiamin Liu}
\author[1]{\small Evrim Turkbey}
\author[1]{\small Ronald M. Summers}
\affil[1]{\small Imaging Biomarkers and Computer-Aided Diagnosis Laboratory,\\
Radiology and Imaging Sciences, National Institutes of Health Clinical Center, Bethesda, MD 20892-1182, USA.}
\maketitle
\begin{abstract}
Automatic organ segmentation is an important yet challenging problem for medical image analysis. The pancreas is an abdominal organ with very high anatomical variability. This inhibits previous segmentation methods from achieving high accuracies, especially compared to other organs such as the liver, heart or kidneys. In this paper, we present a probabilistic bottom-up approach for pancreas segmentation in abdominal computed tomography (CT) scans, using multi-level deep convolutional networks (ConvNets). We propose and evaluate several variations of deep ConvNets in the context of hierarchical, coarse-to-fine classification on image patches and regions, i.e. superpixels. We first present a dense labeling of local image patches via P-ConvNet and nearest neighbor fusion. Then we describe a regional ConvNet (\roneconvnet{}) that samples a set of bounding boxes around each image superpixel at different scales of contexts in a ``zoom-out'' fashion. Our ConvNets learn to assign class probabilities for each superpixel region of being pancreas. Last, we study a stacked \rtwoconvnet{} leveraging the joint space of CT intensities and the \pconvnet{} dense probability maps. Both 3D Gaussian smoothing and 2D conditional random fields are exploited as structured predictions for post-processing. We evaluate on CT images of \Npatients{} patients in 4-fold cross-validation. We achieve a Dice Similarity Coefficient of \trainperf{} in training and \testperf{} in testing.
\end{abstract}
\section{Introduction}
Segmentation of the pancreas can be a prerequisite for computer aided diagnosis (CADx) systems that provide quantitative organ volume analysis, e.g. for diabetic patients. Accurate segmentation could also necessary for computer aided detection (CADe) methods to detect pancreatic cancer. Automatic segmentation of numerous organs in computed tomography (CT) scans is well studied with good performance for organs such as liver, heart or kidneys, where Dice Similarity Coefficients (DSC) of $>$90\% are typically achieved \cite{Wang2014Miccai,Chu2013Miccai,wolz2013automated,Ling08liver}. However, achieving high accuracies in automatic pancreas segmentation is still a challenging task. The pancreas' shape, size and location in the abdomen can vary drastically between patients. Visceral fat around the pancreas can cause large variations in contrast along its boundaries in CT (see Fig. \ref{fig:patch-convnet}). Previous methods report only 46.6\% to 69.1\% DSCs \cite{Wang2014Miccai,Chu2013Miccai,wolz2013automated,farag2014bottom}. Recently, the availability of large annotated datasets and the accessibility of affordable parallel computing resources via GPUs have made it feasible to train deep convolutional networks (ConvNets) for image classification. Great advances in natural image classification have been achieved \cite{krizhevsky2012imagenet}. However, deep ConvNets for semantic image segmentation have not been well studied \cite{mostajabi2014feedforward}. Studies that applied ConvNets to medical imaging applications also show good promise on detection tasks \cite{cirecsan2013mitosis,roth2014new}. In this paper, we extend and exploit ConvNets for a challenging organ segmentation problem. 
\section{Methods}
We present a coarse-to-fine classification scheme with progressive pruning for pancreas segmentation. Compared with previous top-down multi-atlas registration and label fusion methods, our models approach the problem in a bottom-up fashion: from dense labeling of image patches, to regions, and the entire organ. Given an input abdomen CT, an initial set of superpixel regions is generated by a coarse cascade process of random forests based pancreas segmentation as proposed by \cite{farag2014bottom}. These pre-segmented superpixels serve as regional candidates with high sensitivity ($>$97\%) but low precision. The resulting initial DSC is $\sim$27\% on average. Next, we propose and evaluate several variations of ConvNets for segmentation refinement (or pruning). A dense local image patch labeling using an axial-coronal-sagittal viewed patch (\pconvnet{}) is employed in a sliding window manner. This generates a per-location probability response map $P$. A regional ConvNet (\roneconvnet{}) samples a set of bounding boxes covering each image superpixel at multiple spatial scales in a ``zoom-out'' fashion \cite{mostajabi2014feedforward,girshick2014rich} and assigns probabilities of being pancreatic tissue. This means that we not only look at the close-up view of superpixels, but gradually add more contexts to each candidate region. $R_1$-ConvNet operates directly on the CT intensity. Finally, a stacked regional \rtwoconvnet{} is learned to leverage the joint convolutional features of CT intensities and probability maps $\textsl{P}$. Both 3D Gaussian smoothing and 2D conditional random fields for structured prediction are exploited as post-processing. Our methods are evaluated on CT scans of \Npatients{} patients in 4-fold cross-validation (rather than ``leave-one-out'' evaluation \cite{Wang2014Miccai,Chu2013Miccai,wolz2013automated}). We propose several new ConvNet models and advance the current state-of-the-art performance to a DSC of \testmean{} in testing. To the best of our knowledge, this is the highest DSC reported in the literature to date. 
\subsection{Candidate region generation}
\label{sec:region_candidates}
We describe a coarse-to-fine pancreas segmentation method employing multi-level deep ConvNet models. Our hierarchical segmentation method decomposes any input CT into a set of local image superpixels $S=\left\{S_1,\ldots,S_N\right\}$. After evaluation of several image region generation methods \cite{achanta2012slic}, we chose \textit{entropy rate} \cite{liu2011entropy} to extract $N$ superpixels on axial slices. This process is based on the criterion of DSCs given optimal superpixel labels, in part inspired by the PASCAL semantic segmentation challenge \cite{everingham2014pascal}. The optimal superpixel labels achieve a DSC upper-bound and are used for supervised learning below. Next, we use a two-level cascade of random forest (RF) classifiers as in \cite{farag2014bottom}. We only operate the RF labeling at a low class-probability cut $>$0.5 which is sufficient to reject the vast amount of non-pancreas superpixels. This retains a set of superpixels $\{S_\mathrm{RF}\}$ with high recall ($>$97\%) but low precision. After initial candidate generation, over-segmentation is expected and observed with low DSCs of $\sim$27\%. The optimal superpixel labeling is limited by the ability of superpixels to capture the true pancreas boundaries at the per-pixel level with $DSC_\mathrm{max}=80.5\%$, but is still much above previous state-of-the-art \cite{Wang2014Miccai,Chu2013Miccai,wolz2013automated,farag2014bottom}. These superpixel labels are used for assessing `positive' and `negative' superpixel examples for training. Assigning image regions drastically reduces the amount of ConvNet observations needed per CT volume compared to a purely patch-based approach and leads to more balanced training data sets. Our multi-level deep ConvNets will effectively prune the coarse pancreas over-segmentation to increase the final DSC measurements. 
\subsection{Convolutional neural network (ConvNet) setup}
\begin{figure}[htb]
 \centering	
  \resizebox{\figscale\textwidth}{!}{\includegraphics{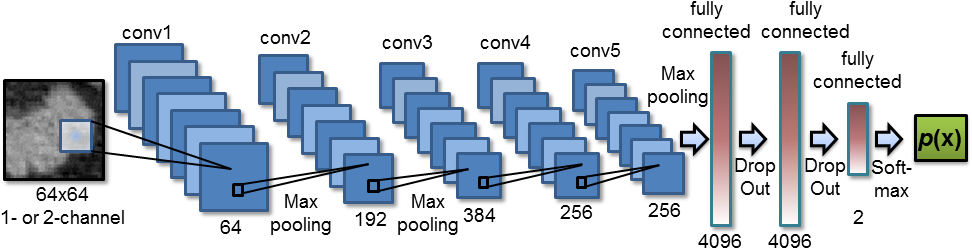}}
	\caption{The proposed ConvNet architecture. The number of convolutional filters and neural network connections for each layer are as shown. This architecture is constant for all ConvNet variations presented in this paper (apart from the number of input channels): \pconvnet{}, \roneconvnet{}, and \rtwoconvnet{}.}
	\label{fig:convnet}
\end{figure}

We use ConvNets with an architecture for binary image classification. Five layers of \textit{convolutional} filters compute and aggregate image features. Other layers of the ConvNets perform \textit{max-pooling} operations or consist of fully-connected neural networks. Our ConvNet ends with a final two-way layer with \textit{softmax} probability for `pancreas' and `non-pancreas' classification (see Fig. \ref{fig:convnet}). The \textit{fully-connected} layers are constrained using ``DropOut'' in order to avoid over-fitting by acting as a regularizer in training \cite{srivastava2014dropout}. GPU acceleration allows efficient training (we use \textit{cuda-convnet2}\footnote{\url{https://code.google.com/p/cuda-convnet2}}). 
\begin{figure}[htb]
\centering	\resizebox{\figscale\textwidth}{!}{\includegraphics{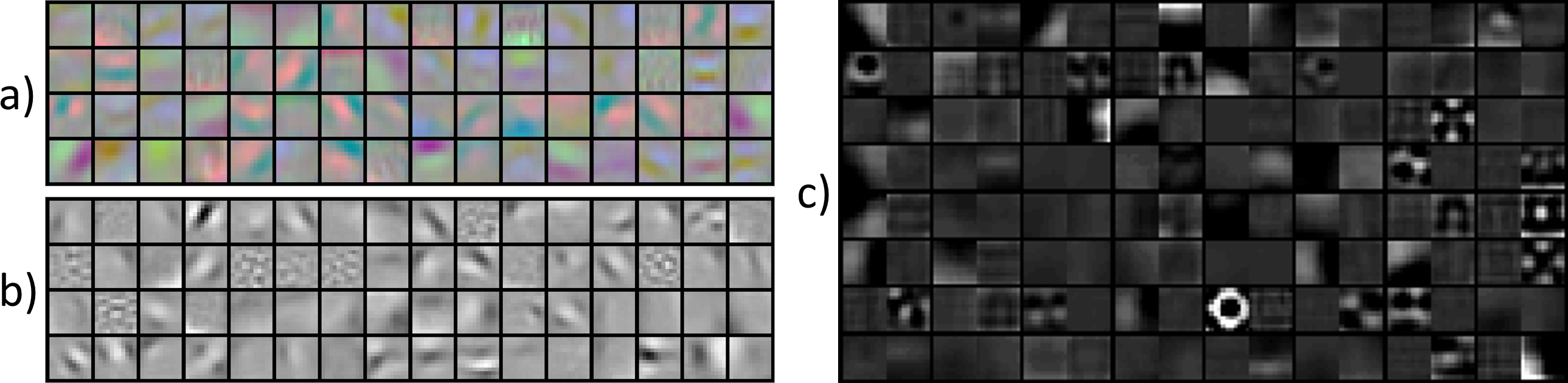}}
	\caption{The first layer of learned convolutional kernels using three representations: a) 2.5D sliding-window patches (\pconvnet{}), b) CT intensity superpixel regions (\roneconvnet{}), and c) CT intensity + $P_0$ map over superpixel regions (\rtwoconvnet{}).}
	\label{fig:conv1}
\end{figure}
\subsection{\pconvnet{}: Deep patch classification}
\begin{figure}[htb]
\centering	\resizebox{\figscale\textwidth}{!}{\includegraphics{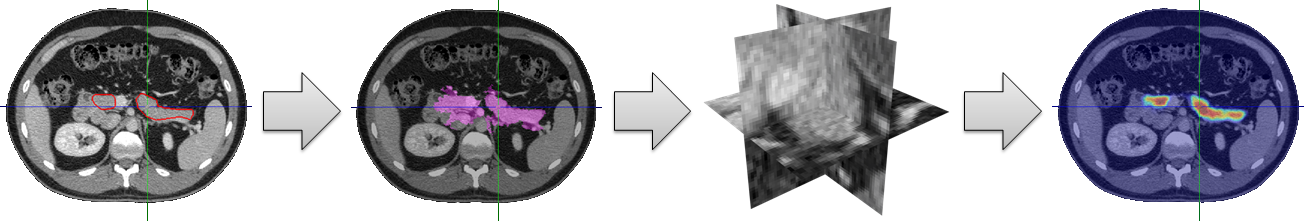}}
	\caption{Axial CT slice of a manual (gold standard) segmentation of the pancreas. From \textit{left} to \textit{right}, there are the ground-truth segmentation  contours (in red), RF based coarse segmentation $\{S_\mathrm{RF}\}$, and the deep patch labeling result using \pconvnet{}.}			
	\label{fig:patch-convnet}
\end{figure}

We use a sliding window approach that extracts 2.5D image patches composed of axial, coronal and sagittal planes within all voxels of the initial set of superpixel regions $\{S_\mathrm{RF}\}$ (see Fig. \ref{fig:patch-convnet}). The resulting ConvNet probabilities are denoted as $P_0$ hereafter. For efficiency reasons, we extract patches every $n$ voxels and then apply nearest neighbor interpolation. This seems sufficient due to the already high quality of $P_0$ and the use of overlapping patches to estimate the values at skipped voxels.
\subsection{\rconvnet{}: Deep region classification} 
\begin{figure}[htb]
\centering	\resizebox{\figscale\textwidth}{!}{\includegraphics{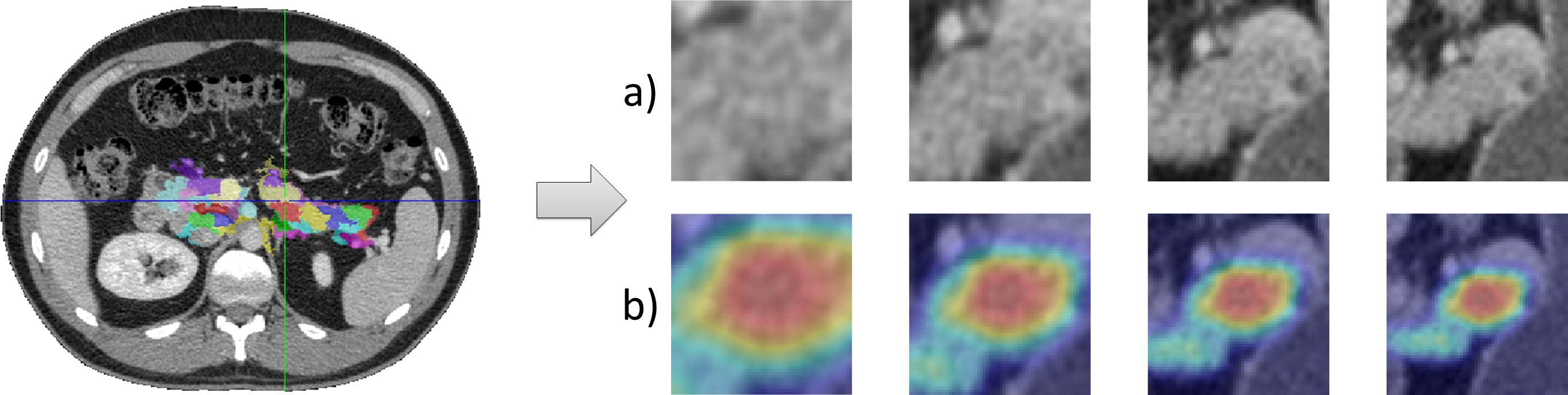}}
	\caption{Region classification using \rconvnet{} at different scales: a) one-channel input based on the intensity image only, and b) two-channel input with additional patch-based \pconvnet{} response.}
	\label{fig:region-convnet}
\end{figure}

We employ the region candidates as inputs. Each superpixel $\in \{S_\mathrm{RF}\}$ will be observed at several scales $N_s$ with an increasing amount of surrounding contexts (see Fig. \ref{fig:region-convnet}). Multi-scale contexts are important to disambiguate the complex anatomy in the abdomen. We explore two approaches: \roneconvnet{} only looks at the CT intensity images extracted from multi-scale superpixel regions, and a stacked \rtwoconvnet{} integrates an additional channel of patch-level response maps $P_0$ for each region as input. As a superpixel can have irregular shapes, we warp each region into a regular square (similar to RCNN \cite{girshick2014rich}) as is required by most ConvNet implementations to date. The ConvNets automatically train their convolutional filter kernels from the available training data. Examples of trained first-layer convolutional filters for \pconvnet{}, \roneconvnet{}, \rtwoconvnet{} are shown in Fig. \ref{fig:conv1}. Deep ConvNets behave as effective image feature extractors that summarize multi-scale image regions for classification.  
\subsection{Data augmentation}
\label{sec:augmentation}
Our ConvNet models (\roneconvnet{}, \rtwoconvnet{}) sample the bounding boxes of each superpixel $\in \{S_\mathrm{RF}\}$ at different scales $s$. During training, we randomly apply non-rigid deformations $t$ to generate more data instances. The degree of deformation is chosen so that the resulting warped images resemble plausible physical variations of the medical images. This approach is commonly referred to as data augmentation and can help avoid over-fitting \cite{krizhevsky2012imagenet,cirecsan2013mitosis}. Each non-rigid training deformation $t$ is computed by fitting a thin-plate-spline (TPS) to a regular grid of 2D control points $\left\{\omega_i;i=1,2,\ldots,k\right\}$. These control points are randomly transformed within the sampling window and a deformed image is generated using a radial basic function $\phi(r)$, where $t(x) = \sum\nolimits^k_{i=1} c_i \phi\left(\left\|x-\omega_i\right\|\right)$ is the transformed location of $x$ and $\{c_{i}\}$ is a set of mapping coefficients.
\subsection{Cross-scale and 3D probability aggregation}
\label{sec:3D_aggregation}
At testing, we evaluate each superpixel at $N_s$ different scales. The probability scores for each superpixel being pancreas are averaged across scales: $p(x) = \frac{1}{N_s } \sum\nolimits_{i=1}^{N_s}p_i(x)$. Then the resulting per-superpixel ConvNet classification values $\{p_1(x)\}$ and $\{p_2(x)\}$ (according to \roneconvnet{} and \rtwoconvnet{}, respectively), are  directly assigned to every pixel or voxel residing within any superpixel $\in \{S_\mathrm{RF}\}$. This process forms two per-voxel probability maps $P_1(x)$ and $P_2(x)$. Subsequently, we perform 3D Gaussian filtering in order to average and smooth the ConvNet probability scores across CT slices and within-slice neighboring regions. 3D isotropic Gaussian filtering can be applied to any $P_k(x)$ with $k=0,1,2$ to form smoothed $G(P_k(x))$. This is a simple way to propagate the 2D slice-based probabilities to 3D by taking local 3D neighborhoods into account. In this paper, we do not work on 3D supervoxels due to computational efficiency\footnote{Supervoxel based regional ConvNets need at least one-order-of-magnitude wider input layers and thus have significantly more parameters to train.} and generality issues. We also explore conditional random fields (CRF) using an additional ConvNet trained between pairs of neighboring superpixels in order to detect the pancreas edge (defined by pairs of superpixels having the same or different object labels). This acts as the \textit{boundary term} together with the \textit{regional term} given by \rtwoconvnet{} in order to perform a min-cut/max-flow segmentation \cite{boykov2006graph}. Here, the CRF is implemented as a 2D graph with connections between directly neighboring superpixels. The CRF weighting coefficient between the boundary and the unary regional term is calibrated by grid-search.
\section{Results \& Discussion}
\subsubsection{Data:}
Manual tracings of the pancreas for \Npatients{} contrast-enhanced abdominal CT volumes were provided by an experienced radiologist. Our experiments are conducted using 4-fold cross-validation in a random hard-split of \Npatients{} patients for training and testing folds with 21, 21, 20, and 20 patients for each testing fold. We report both training and testing segmentation accuracy results. Most previous work \cite{Wang2014Miccai,Chu2013Miccai,wolz2013automated} uses leave-one-patient-out cross-validation protocols which are computationally expensive (e.g., $\sim15$ hours to process one case using a powerful workstation \cite{Wang2014Miccai}) and may not scale up efficiently towards larger patient populations. More patients (i.e. ~20) per testing fold make the results more representative for larger population groups.
\subsubsection{Evaluation:} 
\begin{figure}[t]
\centering	\resizebox{\figscale\textwidth}{!}{\includegraphics{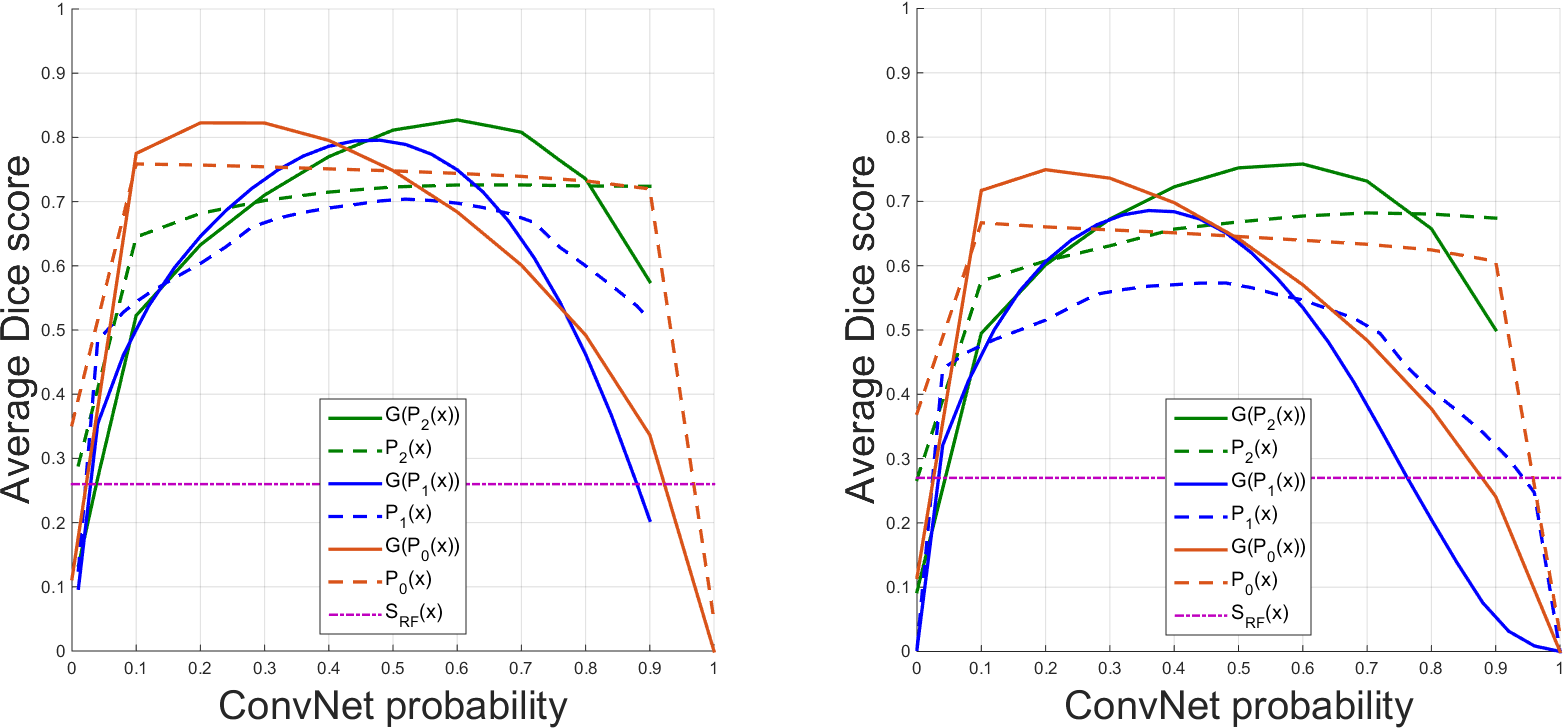}}
	\caption{Average DSCs as a function of un-smoothed $P_k(x), k=0,1,2$, and 3D smoothed $G(P_k(x)), k=0,1,2$, ConvNet probability maps in training (left) and testing (right) in one cross-validation fold}.					
	\label{fig:froc}
\end{figure}

\begin{figure}[htb]
\centering	\resizebox{\figscale\textwidth}{!}{\includegraphics{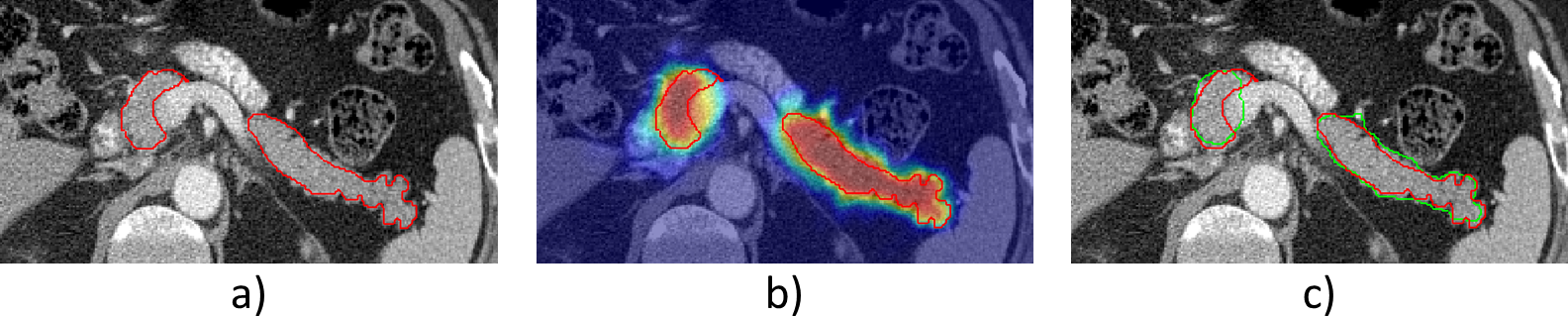}}
	\caption{Example of pancreas segmentation using the proposed \rtwoconvnet{} approach in testing. a) The manual ground truth annotation (in red outline); b) the $G(P_2(x))$ probability map; c) the final segmentation (in green outline) at $p_2=0.6$ (DSC=82.7\%).}
	\label{fig:axial_examples}
\end{figure}

The ground truth superpixel labels are derived as described in Sec. \ref{sec:region_candidates}. The optimally achievable DSC for superpixel classification (if classified perfectly) is 80.5\%. Furthermore, the training data is artificially increased by a factor $N_s\times~N_t$ using the data augmentation approach with both scale and random TPS deformations at the \rconvnet{} level (Sec. \ref{sec:augmentation}). Here, we train on augmented data using $N_s=4$, $N_t=8$. In testing we use $N_s=4$ (without deformation based data augmentation) and $\sigma=3$ voxels (as 3D Gaussian filtering kernel width) to compute smoothed probability maps $G(P(x))$. By tuning our implementation of \cite{farag2014bottom} at a low operating point, the initial superpixel candidate labeling achieves the average DSCs of only 26.1\% in testing; but has a 97\% sensitivity covering all pancreas voxels. Fig. \ref{fig:froc} shows the plots of average DSCs using the proposed ConvNet approaches, as a function of $P_k(x)$ and $G(P_k(x))$ in both training and testing for one fold of cross-validation. Simple Gaussian 3D smoothing (Sec. \ref{sec:3D_aggregation}) markedly improved the average DSCs in all cases. Maximum average DSCs can be observed at $p_0=0.2$, $p_1=0.5$, and $p_2=0.6$ in our training evaluation after 3D Gaussian smoothing for this fold. These calibrated operation points are then fixed and used in testing cross-validation to obtain the results in Table \ref{tab:test_results}. Utilizing \rtwoconvnet{} (stacked on \pconvnet{}) and Gaussian smoothing ($G(P_2(x))$), we achieve a final average DSC of \testmean{}\% in testing, an improvement of 45.7\% compared to the candidate region generation stage at 26.1\%. $G(P_0(x))$ also performs well wiht 69.5\% mean DSC and is more efficient since only dense deep patch labeling is needed. Even though the absolute difference in DSC between $G(P_0(x))$ and $G(P_2(x))$ is small, the surface-to-surface distance improves significantly from 1.46$\pm$1.5mm to 0.94$\pm$0.6mm, (p$<$0.01). An example of pancreas segmentation at this operation point is shown in Fig. \ref{fig:axial_examples}. Training of a typical \rconvnet{} with $N\times~N_s\times N_t=\sim{850k}$ superpixel examples of size $64\times 64$ pixels (after warping) takes $\sim$55 hours for 100 epochs on a modern GPU (Nvidia GTX Titan-Z). However, execution run-time in testing is in the order of only 1 to 3 minutes per CT volume, depending on the number of scales $N_s$. Candidate region generation in Sec. \ref{sec:region_candidates} consumes another 5 minutes per case.

To the best of our knowledge, this work reports the highest average DSC with \testmean{}\% in testing. Note that a direct comparison to previous methods is not possible due to lack of publicly available benchmark datasets. We will share our data and code implementation for future comparisons\footnote{\url{http://www.cc.nih.gov/about/SeniorStaff/ronald_summers.html}}\footnote{\url{http://www.holgerroth.com/}}. Previous state-of-the-art results are at $\sim$68\% to $\sim$69\% \cite{Wang2014Miccai,Chu2013Miccai,wolz2013automated,farag2014bottom}. In particular, DSC drops from 68\% (150 patients) to 58\% (50 patients) under the leave-one-out protocol \cite{wolz2013automated}. Our results are based on a 4-fold cross-validation. The performance degrades gracefully from training (\trainperf{}) to testing (\testperf{}) which demonstrates the good generality of learned deep ConvNets on unseen data. This difference is expected to diminish with more annotated datasets. Our methods also perform with better stability (i.e., comparing \teststd{}\% versus 18.6\% \cite{Wang2014Miccai}, 15.3\% \cite{Chu2013Miccai} in the standard deviation of DSCs). Our maximum test performance is \testmax{}\% DSC with 10\%, 30\%, 50\%, 70\%, 80\%, and 90\% of cases being above 81.4\%, 77.6\%, 74.2\%, 69.4\%, 65.2\% and 58.9\%, respectively. Only 2 outlier cases lie below 40\% DSC (mainly caused by over-segmentation into other organs). The remaining 80 testing cases are all above 50\%. The minimal DSC value of these outliers is \testmin{}\% for $G(P_2(x))$. However \cite{Wang2014Miccai,Chu2013Miccai,wolz2013automated,farag2014bottom} all report gross segmentation failure cases with DSC even below 10\%. Lastly, the variation $CRF(P_2(x))$ of enforcing $P_2(x)$ within a structured prediction CRF model achieves only 68.2\% $\pm$4.1\%. This is probably due to the already high quality of $G(P_0)$ and $G(P_2)$ in comparison.
\begin{table}[htb]
\scriptsize
  \centering
  \caption{\textbf{4-fold cross-validation}: optimally achievable DSCs, our initial candidate region labeling using $S_{RF}$, DSCs on $P(x)$ and using smoothed $G(P(x))$, and a CRF model for structured prediction (best performance in bold).}
  \begin{tabu} to \textwidth {l|c|c|c|c|c|c|c|c|c}
    \toprule
		\toprule
    \textbf{DSC (\%)} & \textbf{Opt.} & $S_{RF(x)}$ & $P_0(x)$ & $G(P_0(x))$ & $P_1(x)$ & $G(P_1(x))$ & $P_2(x)$ & $G(P_2(x))$ & $CRF(P_2(x))$\\
		\midrule
\textbf{Mean} & 80.5 	& 26.1	& 60.9	& 69.5	& 56.8	& 62.9	& 64.9	& \textbf{\testmean}	& 68.2\\
\textbf{Std}  	& 3.6	& 7.1	& 10.4	& 9.3	& 11.4	& 16.1	& 8.1	& \textbf{\teststd}	& 4.1\\
\textbf{Min} 	& 70.9	& 14.2	& 22.9	& 35.3	& 1.3	& 0.0	& 33.1	& \textbf{\testmin}	& 59.6\\
\textbf{Max} 	& 85.9	& 45.8	& 80.1	& 84.4	& 77.4	& 87.3	& 77.9	& \textbf{\testmax}        & 74.2\\
    \bottomrule
		\bottomrule
  \end{tabu}
  \label{tab:test_results}%
\end{table}%
\section{Conclusion}
We present a bottom-up, coarse-to-fine approach for pancreas segmentation in abdominal CT scans. Multi-level deep ConvNets are employed on both image patches and regions. We achieve the highest reported DSCs of \testperf{} in testing and \trainperf{} in training, at the computational cost of a few minutes, not hours as in \cite{Wang2014Miccai,Chu2013Miccai,wolz2013automated}. The proposed approach can be incorporated into multi-organ segmentation frameworks by specifying more tissue types since ConvNet naturally supports multi-class classifications \cite{krizhevsky2012imagenet}. Our deep learning based organ segmentation approach could be generalizable to other segmentation problems with large variations and pathologies, e.g. tumors.
\paragraph{\textbf{Acknowledgments:}} This work was supported by the Intramural Research Program of the NIH Clinical Center. The final publication will be available at Springer.
\bibliographystyle{chicago}
\bibliography{paper397}
\end{document}